\documentclass[acmsmall]{acmart}

\usepackage{setspace}

\usepackage{multirow}
\usepackage{svg}
\usepackage{amsmath}
\usepackage{blindtext}
\usepackage{tcolorbox}
\usepackage{graphicx}
\usepackage{wrapfig}
\usepackage[skip=2pt]{caption} % example skip set to 2pt
\usepackage{subcaption}
\usepackage{latexsym}
\usepackage{graphicx}
\usepackage{adjustbox}
\usepackage{soul}

\usepackage{tikz}
\DeclareMathOperator*{\argmax}{arg\,max}

\usepackage[leftcaption]{sidecap}
\usepackage[linesnumbered,ruled,vlined]{algorithm2e}

\renewcommand\footnotetextcopyrightpermission[1]{}
\usepackage{enumitem}

\AtBeginDocument{%
  \providecommand\BibTeX{{%
    \normalfont B\kern-0.5em{\scshape i\kern-0.25em b}\kern-0.8em\TeX}}}

\begin{document}
\title{\modelname: Self-Supervised Low-Resource Taxonomy Expansion using Large Language Models}

\author{Sahil Mishra}
\affiliation{%
  \institution{Indian Institute of Technology Delhi}
  \city{Delhi}
  \country{India}}
\email{eez238354@ee.iitd.ac.in}

\author{Ujjwal Sudev}
\affiliation{%
  \institution{Samsung Research and Development Institute, Noida}
  \city{Noida}
  \country{India}}
\email{ujjwal.sudev@samsung.com}

\author{Tanmoy Chakraborty}
\affiliation{%
  \institution{Indian Institute of Technology Delhi}
  \city{Delhi}
  \country{India}}
\email{tanchak@iitd.ac.in}

\newcommand{\modelname}{\texttt{FLAME}}

\begin{abstract}
Taxonomies represent an arborescence hierarchical structure that establishes relationships among entities to convey knowledge within a specific domain. Each edge in the taxonomy signifies a hypernym-hyponym relationship. Taxonomies find utility in various real-world applications, such as e-commerce search engines and recommendation systems. Consequently, there arises a necessity to enhance these taxonomies over time. However, manually curating taxonomies with neoteric data presents challenges due to limitations in available human resources and the exponential growth of data. Therefore, it becomes imperative to develop automatic taxonomy expansion methods. Traditional supervised taxonomy expansion approaches encounter difficulties stemming from limited resources, primarily due to the small size of existing taxonomies. This scarcity of training data often leads to overfitting. In this paper, we propose \modelname\ (\textbf{F}ine-tuning \textbf{LA}rge language \textbf{M}odels for taxonomy  \textbf{E}xpansion), a novel approach for taxonomy expansion in low-resource environments (i.e., limited size of existing taxonomies, lack of robust representation capabilities of pre-trained language models, etc.) by harnessing the capabilities of large language models (LLMs) that are trained on extensive real-world knowledge. LLMs help compensate for the scarcity of domain-specific knowledge. Specifically, \modelname\ leverages prompting in few-shot settings to extract the inherent knowledge within the LLMs, ascertaining the hypernym entities within the taxonomy. Furthermore, it employs reinforcement learning to fine-tune the large language models, resulting in more accurate predictions. Experiments on three real-world benchmark datasets demonstrate the effectiveness of \modelname\ in real-world scenarios, achieving a remarkable improvement of 18.5\%  in accuracy and 12.3\% in Wu \& Palmer metric over eight baselines. Furthermore, we elucidate the strengths and weaknesses of \modelname\ through an extensive case study, error analysis and ablation studies on the benchmarks.
\end{abstract}

\begin{CCSXML}
<ccs2012>
<concept>
<concept_id>10010147.10010178.10010179.10003352</concept_id>
<concept_desc>Computing methodologies~Information extraction</concept_desc>
<concept_significance>500</concept_significance>
</concept>
<concept>
<concept_id>10002951.10003317.10003347.10003352</concept_id>
<concept_desc>Information systems~Information extraction</concept_desc>
<concept_significance>300</concept_significance>
</concept>
</ccs2012>
\end{CCSXML}

\ccsdesc[500]{Computing methodologies~Information extraction}
\ccsdesc[300]{Information systems~Information extraction}

\keywords{Taxonomy Expansion, Large Language Models, Self-supervised Learning.}


\maketitle

\section{Introduction}

The term \textit{taxonomy}, in a broader sense, refers to a domain-specific hierarchical knowledge graph which serves as a means to articulate the intricate relationships between terms or concepts, primarily utilizing the semantic relationships like hypernymy, also represented using \textit{"is-a"} relation. The adaptability of taxonomy finds its utility across a diverse array of domains, demonstrating its versatility and importance in academia, e-commerce, health, safety, etc. In academia, educators leverage Bloom's Taxonomy to craft comprehension-based questions \cite{sahubloom21} and quizzes \cite{Elkins2024how}. MeSH \cite{lipscomb2000medical}, Wikidata \cite{vrandevcic2012wikidata}, and DBPedia \cite{fossati2015unsupervised} are employed by researchers to enhance information retrieval systems, enabling more efficient access to relevant knowledge and data.

E-commerce conglomerates such as Amazon \cite{mao2020octet, ziegler2004taxonomy, karamanolakis2020txtract}, eBay \cite{pascalau2010managing}, Pinterest \cite{mahabal2023producing}, and Alibaba \cite{luo2020alicoco} rely on well-structured taxonomies to categorize and display their extensive product inventories, which allows them to make intelligent product recommendations. Even social media platforms like Quora \cite{bolotova2022non} harness taxonomies to suggest related queries, creating a more engaging and informative user experience. Furthermore, risk assessment organizations like Confidential Incident Reporting \& Analysis System (CIRAS) develop taxonomies to identify human-prone errors and accidents, enhancing workplace and industrial safety practices \cite{hale2013working, wallace2016beyond}.

\begin{figure}[!t]
\centering
\includegraphics[width=1.0\textwidth]{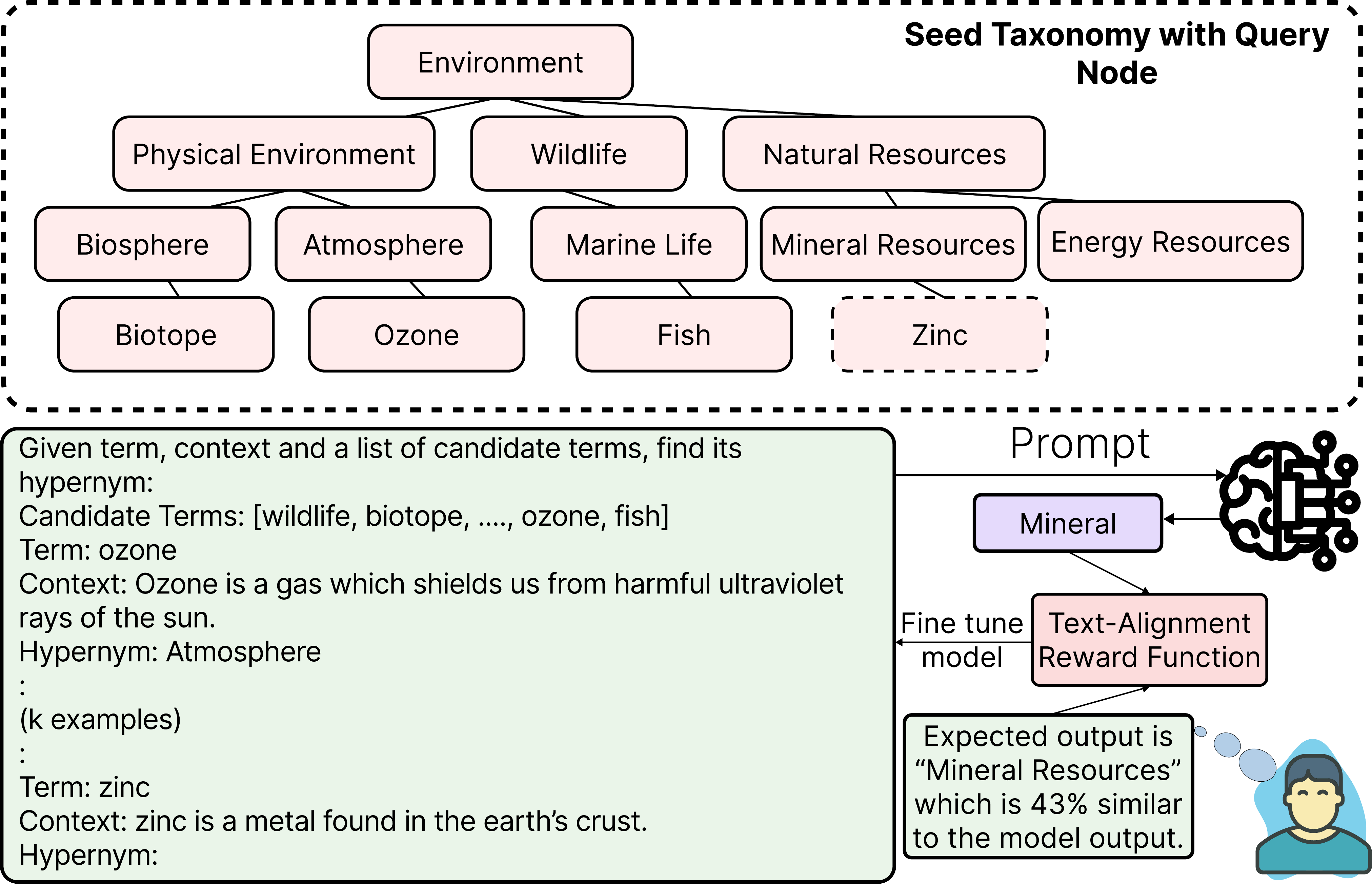}
\caption{A panoramic view of \modelname\ for the taxonomy expansion task to include new concepts such as "\textit{zinc}". A taxonomy-centric prompt is used to fine-tune the LLM (LLaMA-2 7B) using the text-alignment reward function.}
\label{fig:intro}
%\vspace{-5mm}
\end{figure}

These existing taxonomies have been meticulously crafted by domain experts from scratch, requiring substantial effort. Therefore, earlier efforts to automate this process adhered to unsupervised learning paradigms, constructing taxonomies from scratch using hierarchical clustering \cite{zhang2018taxogen} and topic modeling \cite{song2016automatic, wang2013phrase}. However, they not only fail to model the domain intricacies, which are better curated by experts, but also struggle to maintain the coherency of the newly generated taxonomies. Furthermore, as the data grows, new concepts are generated exponentially. This compels us to enhance the existing taxonomies by incorporating these emerging concepts. In this paper, we study the {\em taxonomy expansion} problem -- inserting a query node in the expert-curated taxonomy (alias for seed taxonomy) as a child of another node (also called \textit{anchor node}). Fig. \ref{fig:intro} shows an example of an \textit{environment}-related taxonomy expansion in which a new concept \textit{zinc} has to be inserted.

Recent studies \cite{yu_steam_2020, wang2021enquire, liu2021temp, shen2020taxoexpan, jiang2022taxoenrich} on taxonomy expansion adopt self-supervision, utilizing the available seed taxonomy as a weak signal. They learn the concept semantics by exploiting the structural information of the node present in the seed taxonomy \cite{shen2020taxoexpan, jiang2022taxoenrich, yu_steam_2020, wang2021enquire}, augmented with some external source of information like term definitions \cite{jiang2022taxoenrich, wang2021enquire} or corpora \cite{yu_steam_2020}. This structural information is modeled using ego-nets \cite{shen2020taxoexpan}, local graphs and paths \cite{liu2021temp, wang2021enquire, yu_steam_2020, jiang2022taxoenrich}. A few recent works model the taxonomies using geometric embeddings \cite{jiang2023single}. With the advancement of language models (LMs), another research direction employs prompting to model the node structure and taxonomy summary \cite{xutaxoprompt}. They utilized edges of the seed taxonomy to fine-tune a pre-trained LM \cite{takeoka-etal-2021-low} or perform mask language modeling \cite{xutaxoprompt} to predict the hypernym of a query term. However, all these methods suffer from low-resource challenges. Methods focusing on paths and local graphs solely rely on seed taxonomy for representation learning, which is inherently small. Meanwhile, LM-based methods utilize only lightweight LMs like BERT \cite{devlin_bert_2019} exclusively for hypernymy prediction. Therefore, these methods face limitations due to the lack of robust representation.

To address the challenges of low resources, we propose a novel taxonomy expansion framework, \modelname, which capitalizes on the capabilities of decoder-based large language models (LLMs) to predict hypernyms from the terms present in the seed taxonomy. These LLMs are trained on extensive real-world datasets, inherently imbuing them with a knowledge graph amalgamating various domains, which helps our taxonomy expansion objective. Leveraging few-shot prompting to enable in-context learning (ICL), \modelname\  predicts the hypernyms using a $k$-shot prompt. A $k$-shot ICL \cite{llmarefew} uses \textit{k} labeled samples as demonstrations for the instruction in the prompt to answer the query. Specifically, we model the structure and semantics of the taxonomy using $k$-shot demonstrations. As shown in Fig. \ref{fig:intro}, there is an instruction consisting of a list of candidate terms to predict anchor from, \textit{k} number of examples and a query to predict the hypernym for the term \textit{zinc}. However, these human-engineered prompts are error-prone as the LLMs may not interpret the task description from the prompt in the same manner as a human would. Therefore, following \citet{lester2021power, hu2021lora}, we fine-tune the low-rank parameters of the LLMs to align them with our objective.

Firstly, we perform prompt engineering to design a taxonomy prompt template that emulates the cognitive process of humans while perceiving a taxonomy to retrieve some information from it or expand it. Humans begin with a holistic assessment of the taxonomy to grasp its overall representation. Subsequently, they identify appropriate regions with high confidence in the taxonomy where the query node would be incorporated. Then, they look into finer details to identify the query node's parent, siblings, cousins and uncles. Thus, we prepare the samples from the seed taxonomy for self-supervision and hierarchically cluster them into groups of similar semantics. These semantics are modeled using the definitions used in \cite{wang2021enquire}. Furthermore, we segregate these clusters into sample pools to incorporate local and global neighbourhood knowledge in the prompt, which aids in finding the suitable anchor term for the query term.

Secondly, we design a training strategy aimed at fine-tuning the low-rank parameters of the LLM to align the predicted output with true hypernym. Two predominant strategies to train the LLM to align it with our prompt are supervised fine-tuning (SFT) and reinforcement learning (RL). SFT finetunes the parameters on the next-token prediction objective, while RL aligns the output tokens according to the rules defined by us. We cannot control the token generation in SFT as per our alignment requirements. Therefore, we resort to RL to control only the hypernym tokens generated by the model. Specifically, we train the model using policy gradient optimization. Therefore, we craft a set of lexical and semantic reward signals to guide the policy parameters towards reward maximization through gradient ascent.

Finally, we conduct a series of extensive experiments on three benchmark datasets to compare the efficacy of the \modelname\ framework with eight state-of-the-art (SOTA) baselines. The results show that \modelname\ surpasses the baselines by an average improvement of 18.5\% in accuracy and 12.3\% in Wu \& Palmer score. Ablation studies underscore the importance of few-shot prompting and both global and local instances, contributing to improved model performance. We further discuss case studies and conduct error analysis to illustrate the adeptness of the \modelname\ framework in structural and semantic comprehension.

Our major contributions are summarized as follows\footnote{The source code of \modelname\ is available at \href{https://github.com/sahilmishra0012/FLAME}{https://github.com/sahilmishra0012/FLAME}.}:

\begin{itemize}
    \item We propose an effective taxonomy prompt template to model the structural and semantic representations of nodes present. This template facilitates the retrieval of the hypernymy relationships from LLMs without requiring the complete taxonomy structure.   
    \item We devise a sampling strategy to sample nodes from the taxonomy for the prompt such that they holistically capture both the local intricacies and global neighbourhood of the taxonomy.
    \item We leverage reinforcement learning to fine-tune the low-rank parameters of LLM and align \modelname\ with the specific task of hypernymy prediction.
    \item Extensive experiments and ablations suggest the superiority of \modelname\ over baselines on the taxonomy expansion task, thereby affirming the efficacy of our method.
\end{itemize}

\section{Related Work}
\label{sec:related_work}

% We review two threads of literature: taxonomy construction and prompting.

\subsection{Taxonomy Construction and Expansion}
Earlier methods of taxonomy construction focus on extracting hypernym-hyponym pairs from a corpus and organizing them into a directed acyclic graph (DAG) to build a taxonomy from scratch. They leverage lexical-patterns \cite{hearst1992automatic, nakashole2012patty, snow2004learning, roller2018hearst} to extract hypernymy relationships from the corpus. They also rely on distributional methods \cite{chang2017distributional, fu2014learning, luu2016learning, lin1998information} to generate high dimensional vector representations of the hypernym-hyponym pairs to predict if the hypernymy relationship exists between them. However, some existing taxonomies have already undergone meticulous curation and function in real-time. Therefore, there is a need for solutions for taxonomy expansion. Furthermore, these methods only focus on pairwise relationships, failing to focus on the hierarchies to learn the structural semantics. Therefore, contemporary methods have been developed to expand the existing deployed taxonomies. They follow the self-supervised learning paradigm to capture the structural nuances of the taxonomy using various structural features of the seed taxonomy. They utilize the most common feature, i.e., path. TEMP \cite{liu2021temp} scores on taxonomy-paths using contextual encoders by maximizing the margin loss between positive and negative samples. STEAM \cite{yu_steam_2020} ensembles contextual, lexical and distributional models and utilizes mini-paths to classify hypernymy relationships. TaxoExpan \cite{shen2020taxoexpan} adopts the positional-enhanced graph neural network to represent the anchor node by encoding its ego network. TaxoEnrich \cite{jiang2022taxoenrich} leverages sequential and sibling encoders to encode parent, child and siblings along with the query, scored by the query position matching module. HEF \cite{wang2021enquire} models taxonomy hierarchical structure using pathfinder and stopper to evaluate path and level selections using the fitting score. However, since these methods follow a self-supervised paradigm and mostly use outdated pre-trained encoders, they do not have sufficient resources to learn the taxonomy structure. Therefore, \modelname\ employs LLMs to learn the structural summary from the seed taxonomy and overcomes this low-resource problem to enhance the coherency of the taxonomy.

\subsection{Prompt-based Methods}
Conventional taxonomy expansion methods rely on fine-tuning pre-trained models \cite{liu2021temp}, which suffer from catastrophic forgetting. Meanwhile, the prompting paradigm has gained much attention since the development of LLMs, particularly exemplified by GPT3 \cite{llmarefew}. Prompt engineering techniques \cite{liu2023pre} have been instrumental in enhancing the reasoning abilities of LLMs.  In-context learning \cite{dong2022survey} further complements prompt engineering with the necessary contextual understanding of the instruction to be resolved. Few-shot prompting enhances the prompt by supplying it with the instances elucidating the instruction. Further, soft-prompting techniques \cite{vu2022spot} and parameter-efficient fine-tuning \cite{lester2021power} have been proven pivotal in aligning LLMs with downstream tasks. Recent advancements, such as integrating low-rank decomposition matrices into LLMs \cite{hu2021lora}, have helped reduce the parameters to be fine-tuned for downstream tasks. Recent works on taxonomy expansion involving prompts such as TaxoPrompt \cite{xutaxoprompt} represent the taxonomy summary through a combination of random walk and relation prompts, reformulating the objective to mask language modeling and PLM fine-tuning. However, it encounters low-resource problems even after augmenting self-supervision data through prompting. Our proposed \modelname\ model capitalizes on the few-shot prompting paradigm in conjunction with reinforcement learning to efficiently fine-tune the low-rank parameters of the LLMs.

\section{Problem Formulation}
\label{sec:prob_def}

\textbf{Taxonomy:} A taxonomy $\mathcal{T}^0 = (\mathcal{N}^0, \mathcal{E}^0)$ is a Directed Acyclic Graph (DAG), initiated from a root node representing hypernymy relationship among nodes $\mathcal{N}^0$. Each node $n\in\mathcal{N}^0$ represents a {\em term} with a surface name enriched with the information derived from either a corpus or a definition mined online, either from Wikipedia or generated using LLMs. Each directed edge $\left<n_p, n_c\right>\in\mathcal{E}^0$ signifies an \textit{is-a} or a hypernymy relation directed from the parent or {\em anchor node} $n_p$ towards the child node $n_c$.
\\
\\
\textbf{Taxonomy Expansion with LLMs:}
The taxonomy expansion task refers to the process of augmenting the given {\em seed taxonomy} $\mathcal{T}^0=(\mathcal{N}^0,\mathcal{E}^0)$ with a set of novel concepts $\mathcal{C}$ extracted from a corpus. The expanded taxonomy is represented as $\mathcal{T}=(\mathcal{N},\mathcal{E})=(\mathcal{N}^0\cup\mathcal{C},\mathcal{E}^0\cup\mathcal{R})$, where $\mathcal{R}$ is the new relationship between node ${c} \in \mathcal{C}$ and ${n} \in \mathcal{N}^0$. Specifically,  while inferencing using LLMs, for a \textit{query node} $n_q\in\mathcal{C}$, the model returns a valid \textit{parent node} $n_p$ from the list of anchor nodes $\mathcal{N}^0$. As LLMs generate the text using transition probabilities, they directly return the suitable candidate term based on transition scores. Therefore, there is no provision for defining a scoring metric to rank the list of anchor nodes. As stated in the previous studies \cite{wang2021enquire, shen2020taxoexpan}, relationships among query nodes in $\mathcal{C}$ are disregarded, with each node added independently. The task is illustrated in Figure \ref{fig:intro}.

\section{The {\modelname} Framework}
\label{sec:framework}

\begin{figure*}
    \centering
    \includegraphics[width=1.0\textwidth]{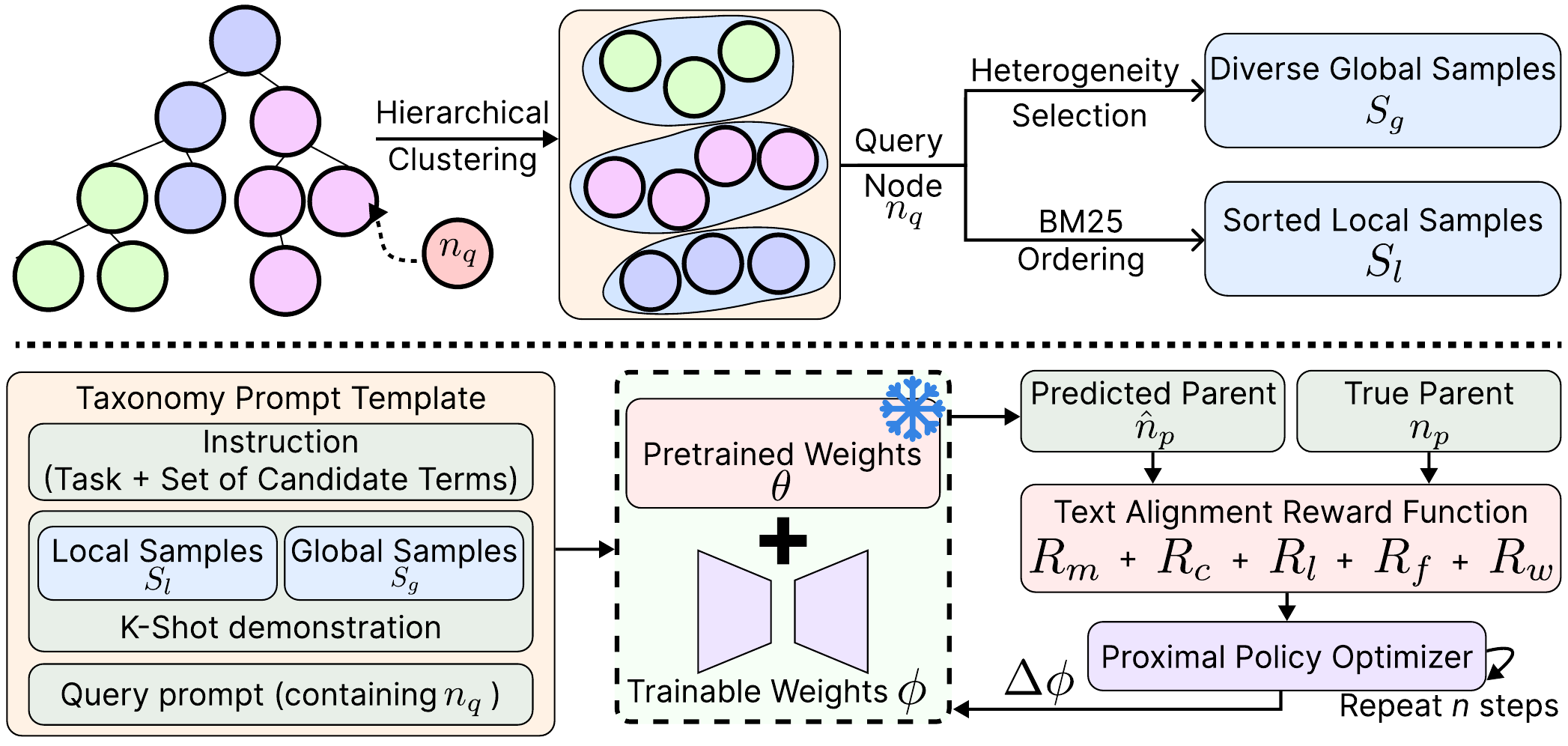}
    \caption{An illustration of \modelname\ framework. Given a seed taxonomy, a $k$-shot prompt is generated (Section \ref{sec:taxo_prompt}) through the taxonomy prompt template (Section \ref{subsec:prompt}) by aggregating nodes into clusters through hierarchical clustering and generating global sample pool $S_g$ (Section \ref{subsec:global}) and local sample pool $S_l$ (Section \ref{subsec:local}).}
    \label{fig:model}
   % \vspace{-5mm}
\end{figure*}

Figure~\ref{fig:model} shows the schematic illustration of our \modelname\ framework. To begin with, we propose a novel taxonomy prompt template based on ICL (discussed in section \ref{sec:taxo_prompt}), which models the local neighbourhood around the query term \footnote{The terms \textit{\textbf{node}} and \textit{\textbf{term}} are used interchangeably.} and maintains a global summary of the taxonomy. Afterwards, leveraging this prompt, we fine-tune the low-rank parameters of the LLM using reinforcement learning (discussed in Section \ref{sec:llm}). Specifically, we employ the proximal policy optimizer to train the model.

\section{Taxonomy Prompt Generation}
\label{sec:taxo_prompt}
Taxonomy is a DAG, which cannot be directly fed into an LLM. Therefore, it needs to be expressed in natural language to facilitate comprehension by the LLM. Thus, there is an imperative need for a prompt function $P(n_q)$, which can directly generate a prompt from the graph corresponding to the query node $n_q$. To achieve this, we first perform hierarchical clustering on the anchor nodes \(\mathcal{N}^0\) in the seed taxonomy, grouping them into $m$ clusters $C_h = \{c_i | 1 \leq i \leq m\}$, where $c_i$ is the cluster with index $i$. We leverage the vector representations of available term definitions to determine centroids and perform clustering.  Further, we generate local and global sample pools to represent the local neighbourhood corresponding to $n_q$ and a generic global overview of the taxonomy.

\subsection{Diverse Global Samples}
\label{subsec:global}

To represent the structural summary of the taxonomy in each prompt, we construct a global sample pool $C_g = \{c_i | n_q \notin c_i\ , c_i \in C_h\}$, which contains the clusters not containing the query node $n_q$ to represent the overall coherency of the taxonomy. The clusters $C_h$ delineate different sub-concepts inherent within the taxonomy. To extract $q$ diverse global samples from $C_g$, we employ heterogeneity selection. We first put the clusters in $C_g$ in the heap by assigning a counter set to zero against them. We then randomly select a node from the cluster with the least count in the heap and increase the cluster count by one (the tie is broken by randomly selecting the cluster). The resultant set of nodes, denoted as $S_g$, constitutes the diverse global sample pool. The detailed procedure to craft $S_g$ is described in Algorithm \ref{algo:diverse}.

\begin{algorithm*}[hb]
\DontPrintSemicolon
\newcommand\mycommfont[1]{\footnotesize\ttfamily{#1}}
\SetCommentSty{mycommfont}

\SetKwInput{KwInput}{Input}                % Set the Input
\SetKwInput{KwOutput}{Output}              % set the Output  
 
  \KwInput{clusters $C_h = \{c_i | 1 \leq i \leq m\}$ \newline query node $n_q$}
  \KwOutput{diverse global sample pool $S_g$}
  $C_g = \{c_i | n_q \notin c_i\ , c_i \in C_h\}$\\
\ForEach{cluster $c \in C_g$}{push $c$ to heap and initialize a zero counter against $c$}
    \Repeat{$q$ times}{
        Pop the lowest count cluster $c_g$ from heap\\
        Randomly select a node $n_g$ from the cluster\\
        Append $n_g$ to $S_g$\\
        Increase the counter against $c_g$ by 1\\
        Push $c_g$ back into the heap
    }

\caption{Diverse Global Sample Pool Generation}
\label{algo:diverse}
\end{algorithm*}

\subsection{Sorted Local Samples}
\label{subsec:local}
Since the task of link prediction requires comprehension of the local neighbourhood graph structure \cite{veličković2018graph}, we must represent the local graph surrounding the link in the prompt. Therefore, we identify the terms akin to the query term $n_q$. In order to articulate the local structure in textual form, we opt for a cluster $C_l = \{c_i | n_q \in c_i \cap c_i \in C_h \}$ from the given set $C_h$ that exhibits the highest semantic similarity to the query node $n_q$ by computing its distance from the centroid of all the clusters. We rank these nodes corresponding to $n_q$ using their BM25 scores \cite{robertson1994some}, resulting in a sorted local samples set $S_l$. BM25 score evaluates the correlation between a node and $n_q$ using their definitions.

\subsection{Taxonomy Prompt Template Engineering}
\label{subsec:prompt}

We construct a taxonomy prompt template to represent the global structure and local intricacies using the sample pools $S_g$ and $S_l$, respectively. As ICL is an important step in prompt engineering \cite{zhao2021calibrate}, we employ a few-shot prompting strategy, extracting pertinent information from these samples, such as their hypernyms and definitions and culminating them into instances $(s_g, d_g, p_g, s_l, d_l, p_l)$, where $d_g$ and $d_l$ are the definitions of the terms $s_g \in S_g$ and $s_l \in S_l$, respectively. We alternately select \textit{k} instances from diverse global samples and sort local samples to formulate a $k$-shot prompt. To construct the query for the prompt, we use the query node $n_q$ and its corresponding definition $d_q$. As our objective is to prompt the model to forecast the parent term or hypernym for $n_q$, we append the set of anchor nodes $\mathcal{N}^0$ from taxonomy in the prompt to the instruction $\mathcal{I}$. These candidate terms encapsulate the whole taxonomy, thereby enriching the prompt with its holistic structure. Therefore, the final prompt has an instruction, a set of anchor nodes, $k$-shot examples and the query $(n_q, d_q)$ awaiting completion. A comprehensive discussion of the prompt template, supplemented with illustrative examples, is provided in Appendix \ref{app:prompt}. The prompt function template is defined as follows:
\begin{equation}
    P(n_q) = \mathcal{I} \oplus \mathcal{N}^0 \oplus_{i=1}^{k} (s_g, d_g, p_g, s_l, d_l, p_l) \oplus (n_q, d_q),
   \label{eq:prompt}
\end{equation}
where \(\oplus\) is the concatenation operation, while \(\oplus_{i=1}^{k}\) represents concatenation of alternately selected {\em k} instances. The prompt template is discussed in Appendix \ref{app:prompt}.

\section{LLM Fine-tuning}
\label{sec:llm}

Given an LLM parameterized by \(\theta\), a $k$-shot prompt denoted by \( P(n_q)\) and the set of possible candidate terms $\mathcal{N}^0$ as a sequence of tokens, a standard $k$-shot prompting on LLM can be elucidated as, 
\begin{equation}
\small
    \hat{n}_p = \argmax_{n_a \in \mathcal{N}^0} p_\theta (n_a \mid P(n_q)),
\end{equation}
where \(\hat{n}_p\) is the predicted parent node. With an ICL prompting, the LLM orchestrates generating a sequence of tokens, predicting the candidate term while concurrently providing an extra set of hypernym-hyponym pairs. The modified process can be delineated as follows,

\begin{equation}
\small
    \hat{n}_p, \hat{a}_p, \hat{a}_c = \argmax_{n_a \in \mathcal{N}^0, a_p, a_c \notin \mathcal{N}^0} p_\theta (n_a,a_p, a_c \mid P(n_q)),
\end{equation}
where $\hat{a}_p$ and $\hat{a}_c$ represent the novel concepts that are not already present in the taxonomy. These concepts form a hypernymy relationship $\hat{a}_p \rightarrow \hat{a}_c$ where $\hat{a}_p$ and $\hat{a}_c$ are the hypernym-hyponym pairs (an example is described in Appendix \ref{app:aug}). Nevertheless, since these concepts are novel, assessing the integrity of these relations poses a challenge as the hypernyms may not exist in the seed taxonomy. Therefore, we disregard these edges.

\subsection{Structure Learning}
\label{subsec:struct_learn}

Quintessential LLMs typically grapple to grasp the taxonomy structure necessary to accurately predict the parent node \(a_p\) from the given set of anchor terms \(\mathcal{N}^0\). Instead, they rely on their existing knowledge to generate tokens semantically similar to the potential candidate term but not the exact one. Also, sometimes, they start completing the prompt with reasoning, which we do not want. Therefore, we aim to align the LLMs to our prompt and restrict their output strictly to the set of given anchor terms. To achieve this, we fine-tune the LLM $\theta$ for the link prediction task. Instead of fine-tuning all the parameters of \(\theta\), we adopt a parameter-efficient fine-tuning approach \cite{lester2021power}, utilizing LoRA adapters \cite{hu2021lora}. For simplicity, we denote the adapter-augmented LM as \(\phi\), clarifying that only the adapter weights are subject to update while training.

To fine-tune LoRA adapters, the two generic approaches are employed, each attending to specific aspects of the prompt. The first method, based on supervised fine-tuning, trains the LLM to minimize the perplexity during the next token generation task on the prompt. This approach, however, does not singularly prioritize the alignment of the LLM with the provided target label. Conversely, we propose an alternative method, following the reinforcement learning paradigms, to train the LLM to precisely synergize with the target label, the parent node, rather than the entire prompt. To fine-tune the LLM \(\phi\) and optimize the objective, we employ policy gradient optimization \cite{wang2022policy}, formalizing the setup as follows.

\subsubsection{Environment}
We resort to a model-free environment adopting a policy network \(\pi_\theta\), which has the LLM \(\phi\) as its core component. Therefore, we adjust the parameters of LLM, observe the differences in resulting rewards, and update $\phi$ towards maximizing reward acquisition. The generated label \(\hat{n}_p\) is defined as the observation of the environment, which is utilized to compute the rewards.

\subsubsection{Action and State Space}
A policy is a mapping from the state space to the action space, meaning the action is decided based on the observation state of the environment. Therefore, we define the prompt \(P(n_q)\), which is fed into the model as the initial state $s_0 \in \mathcal{S}$, where \(\mathcal{S}\) is the state space of the environment. The action space is the vocabulary \(\mathcal{V}\) of the LLM, as the action of this environment is the generation of the token after getting the response from the state. Action at timestep \(t\) is defined as generating a new token at time \(t\). The state at timestep \(t\) is defined as $s_t=\left(s_{t-1},\left\{a_{t-1}\right\}\right)$, indicating the tokens generated till \(t-1\).

\subsubsection{Gradient-Based Policy Optimization}
The parameters $\phi$ governing the policy network \(\pi_\phi : \mathcal{S} \rightarrow \mathcal{V}\) are generally learnt using gradient-based algorithms, which estimate the gradients with respect to cumulative reward $G_t$ at time $t$ to maximize it. A known gradient estimator is stated below:
\begin{equation}
\hat{g}=\hat{\mathbb{E}}_t\left[\nabla_\theta \log \pi_\theta\left(a_t \mid s_t\right) \hat{A}_t\right],
\end{equation}
where $E_t$ is the empirical average over a batch while $A_t$ is an estimator of the advantage function at timestep $t$.

We adhere to the Proximal Policy Optimization (PPO), which has proven to be a great assist in training GPT-3.5 to train the policy $\pi_\phi$. PPO aims to make the policy more likely to select actions with a high advantage value, meaning that actions have a much higher measured cumulative reward than the value function could predict. The advantage function in PPO is defined as:
\begin{equation}
\begin{aligned}
 \hat{A}_t= &\sum_{i=0}^{T-t+1}(\gamma \lambda)^i\delta_{t+i},
\end{aligned}
\end{equation}
 where $\delta_{t+i} = \left[r_{t+i}+\gamma V\left(s_{t+i+1}\right)-V\left(s_{t+i}\right)\right]$, \(r_t\) is the reward at timestep \(t\), \(V(s_t)\) is the value function to compute reward for state \(s_t\) while \(\gamma\) and \(\lambda\) are hyperparameters. \(T\) is the timestep defined for the end of the episode. The policy network works in close association with value function \(V(s_t):\mathcal{S}\xrightarrow{}\mathbb{R}\), which is a feed-forward neural network initialized from the uniform distribution.

It has been observed that PPO exhibits instability within continuous action spaces due to the attenuation of the rewards beyond certain thresholds. It also gets entrenched in sub-optimal actions on discrete action spaces characterized by sparse and elevated rewards. Therefore, to mitigate these issues, we use KL-divergence regularization, inspired by Trust Region Policy Optimization (TRPO) \cite{pmlrschulman15}. KL-divergence measures the distance between a reference model and the policy, ensuring that the new policy is not too different from the reference model after updates. The reference model in this context can be the same as the base model $\phi$. The KL divergence is defined as follows:
\begin{equation}
\mathcal{KL} \left(\pi_\phi \| \pi_{\text{ref}}\right)=\sum_{x \in \mathcal{X}} \pi_\phi(x) \log \left(\frac{\pi_\phi(x)}{\pi_{\text{ref}}(x)}\right),
\end{equation}
where $\pi_{\text{ref}}$ is the reference model. Therefore, the PPO objective, combined with KL-divergence, is defined as follows:
\begin{equation}
\begin{aligned}
& \underset{\phi}{\argmax} \hat{\mathbb{E}}_t\left[\frac{\pi_\phi\left(a_t \mid s_t\right)}{\pi_{\text {ref }}\left(a_t \mid s_t\right)} \hat{A}_t-\beta \mathcal{KL}_t\right],
\end{aligned}
\end{equation}
where  $\mathcal{KL}_t = \mathcal{KL}\left(\pi_{\text {ref }}\left(\cdot \mid s_t\right)\| \pi_\phi\left(\cdot \mid s_t\right)\right)$, and $\mathcal{KL}_t$ is the KL-divergence between the policy $\pi_\phi$ and reference model $\pi_{\text {ref }}$ for state $s$ at time $t$.

\subsubsection{Reward Function}
\label{subsubsec:reward}
The reward function \(r_t\) determines the rewards for the action taken at time \(t\). We define the reward function as the summation of six rewards as,
\begin{equation}
    r_t = R_m + R_c + R_l + R_w + R_f,
\end{equation}
where \(R_m\), \(R_c\), \(R_l\), \(R_f\) and \(R_w\) are lexical and semantic rewards which are computed on true parent term \(n_p\) and predicted parent term \(\hat{n}_p\). They are defined below:
\\
\textbf{Label Reliability:} This reward determines whether a predicted parent is the same as the true parent. It is defined as:
\begin{equation}
R_m=I(\hat{n}_p=n_p)
\end{equation}
\\
\textbf{Semantic Consistency:} This reward determines how much the predicted parent term is semantically similar to the true parent. It is computed as:
\begin{equation}
R_c=\text{cos-sim}(\hat{e},e),
\end{equation}
where \(\hat{e}\) and \(e\) are the vector representations of \(\hat{n}_p\) and \(n_p\), respectively and \(\text{cos-sim}\) is the cosine similarity between vector representations.
\\
\textbf{Label Length Conformity:} This reward checks if the length \(\hat{n}_p\) is same as the length of \(n_p\). It is defined as:
\begin{equation}
\begin{aligned}
    R_l = \Biggl\{\begin{array}{lr}
        -\frac{\big|\mathcal{L}_{\hat{n}_p}-\mathcal{L}_{n_p}\big|}{\mathcal{L}_{\hat{n}_p}+\mathcal{L}_{n_p}}, & \text{if } \big|\mathcal{L}_{\hat{n}_p}-\mathcal{L}_{n_p}\big| \ne 0,\\
        0, & \text{otherwise},
        \end{array}
  \end{aligned}
\end{equation}
where \(\mathcal{L}_{\hat{n}_p}\) and \(\mathcal{L}_{n_p}\) are the lengths of \(\hat{n}_p\) and \(n_p\), respectively, and \(\big|\cdot\big|\) represents the absolute operator.
\\
\textbf{Token Count Consistency:} This reward computes the difference in the count of tokens generated by the model and the count of tokens in the true label.
\begin{equation}
\begin{aligned}
    R_w = \Biggl\{\begin{array}{lr}
        \frac{\big|\mathcal{S}_{\hat{n}_p}\cap\mathcal{S}_{n_p}\big|}{\big|\mathcal{S}_{\hat{n}_p}\cup\mathcal{S}_{n_p}\big|}, & \text{if } \big|\mathcal{S}_{\hat{n}_p}\cap\mathcal{S}_{n_p}\big| \ne 0,\\
        0, & \text{otherwise},
        \end{array}
  \end{aligned}
\end{equation}
where \(\mathcal{S}_{\hat{n}_p}\) and \(\mathcal{S}_{n_p}\) are the sets of tokens present in \(\hat{n}_p\) and \(n_p\) respectively, while \(\big|.\big|\) represents the set cardinality.
\\
\textbf{Fuzzy Label Matching:} The tokens generated by LLM,  despite being prompt containing the set of candidate terms, do not align with the true parent term \(n_p\). However, they exhibit a degree of semantic similarity to \(n_p\). Therefore, we establish this reward system based on approximate or fuzzy string matching by employing the Levenshtein distance metric to assess the level of syntactic similarity between the strings. It is defined as:
\begin{equation}
    R_f = F_r(\hat{n}_p,n_p) + F_{pr}(\hat{n}_p,n_p)+
    F_{tsor}(\hat{n}_p,n_p) + F_{tser}(\hat{n}_p,n_p),
\end{equation}
where \(F_r\) is the edit distance ratio, \(F_{pr}\) is partial edit distance ratio, \(F_{tsor}\) is token sort ratio while \(F_{tser}\) is token set ratio between \(\hat{n}_p\) and \(n_p\). These fuzzy string-matching methodologies are discussed with illustrated examples in Appendix \ref{app:reward}.

Therefore, we optimize the following objective:
\begin{equation}
\min _\phi\left[-\log \left(p_\phi\left(\hat{n}_p \mid P\left(n_q\right)\right)\right)\right]
\end{equation}

\subsection{Inference with LLM}
\label{subsec:infer_llm}
During the inference process, our objective is to predict the hypernym or the parent node \(n_p\) from the given set of anchor nodes \(\mathcal{N}^0\). For each query term \(c \in \mathcal{C}\), we procure its respective definition \(d_c\), obtain both its local and global samples, and construct a prompt akin to the one we construct in Section \ref{subsec:prompt}. Since LLMs do not provide any scoring metric while ranking, \modelname\ directly returns the hypernym of the query term without any score. Due to our fine-tuning of the LLM based on the reward function, it is constrained to exclusively generating the hypernym of the query node, with no additional tokens being generated. Consequently, the generation of new hypernymy relationships such as $\hat{a}_p$--$\hat{a}_c$ does not occur.

\subsection{Complexity Analysis}
\label{subsec:comp_analysis}
The inference time complexity of an LLM is \(\mathcal{O}(\Phi \cdot l_{avg}^2 \cdot d)\), where \(\Phi\) and \(d\) represent the number of parameters and the hidden dimension of the LLM, respectively, while \(l\) represents the average length of the input sequence. Therefore, the time complexity of \modelname\ for a query term during inference is \(\mathcal{O}(|\mathcal{C}| \cdot \Phi \cdot l_{avg}^2 \cdot d)\), which is joint fastest with TaxoPrompt \cite{xutaxoprompt} while the fastest among all the other existing SOTA baselines with time complexity of \(\mathcal{O}(|\mathcal{N}^0| \cdot |\mathcal{C}| \cdot \Phi \cdot l_{avg}^2 \cdot d)\), as LLMs directly generate candidate terms without the need for explicit ranking.

\section{Experimental Setup}
\label{sec:experiment_setup}
This section furnishes a comprehensive overview of the benchmark datasets, baselines for comparative analysis, and evaluation metrics.

\begin{table}[]
\caption{Statistics of the three SemEval-2016 datasets -- Environment (Env), Science (Sci) and Food.  $|\mathcal{N}^0|$ and $|\mathcal{E}^0|$ are the numbers
of nodes and edges in the seed taxonomy, respectively, while $|D|$ is the depth of the taxonomy.}
\label{table:dataset}
\begin{tabular}{lrrr}
\hline
Dataset      & Env & Sci & Food \\ \hline
$|\mathcal{N}^0|$          & 261         & 429     & 1486 \\
$|\mathcal{E}^0|$          & 261         & 452     & 1576\\
$|D|$          & 6           & 8       & 8    \\ \hline
\end{tabular}
%\vspace{-5mm}
\end{table}

\subsection{ Benchmark Datasets}
We use three publicly available benchmark datasets sourced from SemEval-2016 Task 13 Taxonomy Extraction and Evaluation \cite{bordea-etal-2016-semeval}, namely Environment (SemEval-Env), Science (SemEval-Sci) and Food (SemEval-Food) (shown in Table \ref{table:dataset}) in alignment with the existing studies \cite{liu2021temp, xutaxoprompt, wang2021enquire, jiang2023single}. Some baselines like \cite{yu_steam_2020} use the external corpus to coherently model the semantics of the taxonomy. Apart from this, \modelname\, as well as some baselines, utilize definitions of the concepts to achieve similar objective. Therefore, we acquire these corpora and definitions from \citet{yu_steam_2020, jiang2023single, wang2021enquire} and \citet{liu2021temp}. Some definitions are incomplete, mislabeled or corrupted while acquiring the data. We rectify these discrepancies through cross-referencing with Wikipedia and GPT-3.5. 

Several of these benchmark datasets have multiple parent nodes associated with a single node. Therefore, we utilize the spanning tree of the dataset, serving as the seed taxonomy aligning with the problem's definition. Spanning requires the pruning of only 5\% of the edges. Baseline methods \cite{yu_steam_2020,wang2021enquire,shen2020taxoexpan}, similar to \modelname, perform the task of taxonomy expansion by incorporating new terms into the existing seed taxonomy without altering its structure. Therefore, we selectively sample the test set from the leaf nodes to ensure that parent nodes of the test set exist in the seed taxonomy. Consistent with \cite{yu_steam_2020}, we randomly sample 20\% leaf nodes. The remaining nodes constitute a comprehensive seed taxonomy utilized as self-supervision data during training. Even after pruning and train-test split, there remains ample data for training, given that \modelname\ is tailored for scenarios with limited resources.

\subsection{Baseline Methods} 
We compare the performance of \modelname\ framework with eight baseline methods. 
\begin{itemize}[leftmargin=*]
  \item \textbf{BERT+MLP} \cite{devlin_bert_2019} utilizes pre-trained embeddings of the surface names from BERT and leverages a Multi-layer Perceptron for hypernymy detection.
  \item \textbf{TAXI} \cite{panchenko2016taxi}, the winner of SemEval-2016 Task 13, detects hypernym relations through substring matching and pattern extraction.
  \item \textbf{TaxoExpan} \cite{shen2020taxoexpan} models the anchor representation by encoding its ego-net using position-enhanced GNN and scores the parental relationship using a log-bilinear feed-forward model.
  \item \textbf{STEAM} \cite{yu_steam_2020} takes graph-based, contextual, and manually crafted lexical-syntactic features into account for query-anchor pairs and trains via multi-view co-training.
  \item \textbf{TEMP} \cite{liu2021temp} utilizes taxonomic paths from the root node to learn the node representation of the query node via dynamic margin loss.
  \item \textbf{HEF} \cite{wang2021enquire} computes ego tree representation by concatenating embeddings and utilizes fitting score based on Pathfinder and Stopper modules to score the parent nodes.
  \item \textbf{BoxTaxo} \cite{jiang2023single} introduces box embeddings and trains geometric and probabilistic losses to score the parents for query nodes using the volume of the hyper-rectangles.
  \item \textbf{TaxoPrompt} \cite{xutaxoprompt} formulates a prompt template that represents taxonomy context using random walk and optimizes on masked language modeling objective.
\end{itemize}

\subsection{Implementation Details}
We implement the baselines, except BERT+MLP, using the code obtained from their respective code repositories maintained by the authors. The codebase of the \modelname\ framework is developed leveraging Huggingface libraries like PEFT and LoRA in PyTorch. We implement the PPO algorithm using the TRL library. We use 2x80GB NVIDIA A100 GPUs to train the model and 1x80GB NVIDIA A100 GPU for inference. Hyperparameters are tuned on a 30\% subset of the SemEval-Env dataset. The values of the tuned hyperparameters are stated as follows. We use LLaMA-2 7 billion \cite{touvron2023llama} as the core LLM \(\theta\). We integrate low-rank parameters with an attention dimensionality (rank) of 8 and scaling factor \(\alpha\) set at 16 to the LLM \(\theta\), making it \(\phi\). For the RL stage, we use a batch size of 32 and a dropout rate of 0.4 to reduce overfitting while training. The optimizer used is Adafactor \cite{shazeer2018adafactor} with a learning rate of 1.41e-5. The hyperparameters \(\gamma\) and \(\lambda\) are both set to 1.  For inference, we use the following generation parameters: temperature = 0.95, top\_p=1.0, pad\_token\_id = eos\_token\_id, do\_sample = True, num\_beams = 1, max\_length = 1024. We do not define any stopping criteria, as the reward function implicitly restricts LLM from producing extra tokens apart from hypernyms.

\subsection{Evaluation Metrics}
For each query node, some baselines score and rank the candidate terms in the existing taxonomy, while a few directly predict the candidate term. Given the query set \(\mathcal{C}\), let the predictions made by the model be \(\left\{\hat{y}_1, \hat{y}_2, \cdots, \hat{y}_{|\mathcal{C}|}\right\}\) and the ground truth positions be \(\left\{\hat{y}_1, \hat{y}_2, \cdots, \hat{y}_{|\mathcal{C}|}\right\}\). Following \citet{jiang2023single, liu2021temp, manzoor2020expanding, vedula_enriching_2018} and \citet{yu_steam_2020}, we use the following metrics to evaluate the performance of the competing models.
\begin{itemize}
    \item \textbf{Accuracy (Acc):} It measures the count of predicted candidate positions matching the ground-truth positions, calculated as,
        \begin{equation}
            \text{Acc}=\text{Hit@1}=\frac{1}{|\mathcal{C}|}\sum_{i=1}^{|\mathcal{C}|}{\mathbb{I}\left(y_i=\hat{y_i}\right)},
        \end{equation}
    where \(\mathbb{I}(\cdot)\) represents the indicator function.
    % \item \textbf{Mean Reciprocal Rank (MRR):} It computes the average reciprocal rank of the query node's true hypernym among all the predicted candidate terms. Models like TAXI \cite{panchenko2016taxi} and \modelname\ cannot utilize this metric as they do not rank the candidate terms. It is computed as,
    %     \begin{equation}
    %     \text{MRR}=\frac{1}{|\mathcal{C}|}\sum_{i=1}^{|\mathcal{C}|}{\frac{1}{\text{rank}\left(y_i\right)}}
    %     \end{equation}
    \item \textbf{Wu \& Palmer Similarity  (Wu\&P) \cite{wu_verbs_1994}:} It measures the structural similarity between predicted hypernym and ground truth hypernym in the seed taxonomy. It is calculated using the closest common ancestor as,
        \begin{equation}
        \text{Wu\&P}=\frac{1}{|\mathcal{C}|}\sum_{i=1}^{|\mathcal{C}|}{\frac{2\times \text{depth}\left(\text{LCA}\left(\hat{a_i}, a_i\right)\right)}{\text{depth}\left(\hat{a_i}\right)+\text{depth}\left(a_i\right)}},
        \end{equation}
        where $\text{depth}(\cdot)$ denotes the node's depth in the seed taxonomy, and $\text{LCA}(\cdot,\cdot)$ is the least common ancestor of two nodes.

\end{itemize}

Rank-based metrics like Mean Reciprocal Rank (MRR) do not apply to our method as the prompt is designed to predict the parent node directly rather than ranking all anchor nodes.

\begin{table}[t]
\caption{Comparison of \modelname\ with the baseline methods. The mean performance ($\pm$ 1-std dev) of each competing method across three runs with three different seeds is reported in percentage (\%). The best performance is marked in bold, while the best baseline is underlined.}
\scalebox{1.02}{
\begin{tabular}{|l|cc|cc|cc|}
\hline
\multicolumn{1}{|c|}{\multirow{2}{*}{\textbf{Methods}}} & \multicolumn{2}{c|}{\textbf{SemEval16-Env}}                            & \multicolumn{2}{c|}{\textbf{SemEval16-Sci}}                            & \multicolumn{2}{c|}{\textbf{SemEval16-Food}}                           \\ \cline{2-7} 
\multicolumn{1}{|c|}{}                                  & \multicolumn{1}{c|}{\textbf{Acc}}            & \textbf{Wu\&P}          & \multicolumn{1}{c|}{\textbf{Acc}}            & \textbf{Wu\&P}          & \multicolumn{1}{c|}{\textbf{Acc}}            & \textbf{Wu\&P}          \\ \hline
BERT+MLP                                                & \multicolumn{1}{c|}{$12.6 \pm 1.1$}          & $48.3 \pm 0.8$          & \multicolumn{1}{c|}{$12.2 \pm 1.7$}          & $45.1 \pm 1.1$          & \multicolumn{1}{c|}{$12.7 \pm 1.8$}          & $49.1 \pm 1.2$          \\
TAXI                                                    & \multicolumn{1}{c|}{$18.5 \pm 1.3$}          & $47.7 \pm 0.4$          & \multicolumn{1}{c|}{$13.8 \pm 1.4$}          & $33.1 \pm 0.7$          & \multicolumn{1}{c|}{$20.9 \pm 1.1$}          & $41.6 \pm 0.2$          \\
TaxoExpan                                               & \multicolumn{1}{c|}{$10.7 \pm 4.1$}          & $48.5 \pm 1.7$          & \multicolumn{1}{c|}{$24.2 \pm 5.4$}          & $55.6 \pm 1.9$          & \multicolumn{1}{c|}{$24.6 \pm 4.7$}          & $52.6 \pm 2.2$          \\
STEAM                                                   & \multicolumn{1}{c|}{$34.1 \pm 3.4$}          & $65.2 \pm 1.4$          & \multicolumn{1}{c|}{$34.8 \pm 4.5$}          & $72.1 \pm 1.7$          & \multicolumn{1}{c|}{$31.8 \pm 4.3$}          & $64.8 \pm 1.2$          \\
TEMP                                                    & \multicolumn{1}{c|}{$45.5 \pm 8.6$}          & $77.3 \pm 2.8$          & \multicolumn{1}{c|}{$43.5 \pm 7.8$}          & $76.3 \pm 1.5$          & \multicolumn{1}{c|}{$44.5 \pm 0.3$}          &  $\underline{77.2 \pm 1.4}$    \\
HEF                                                     & \multicolumn{1}{c|}{$51.4 \pm 2.8$}          & $71.4 \pm 2.3$          & \multicolumn{1}{c|}{$48.6 \pm 5.3$}          & $72.8 \pm 1.8$          & \multicolumn{1}{c|}{$46.1 \pm 4.3$}          & $73.5 \pm 3.2$          \\
BoxTaxo                                                 & \multicolumn{1}{c|}{$32.3 \pm 5.8$}          & $73.1 \pm 1.2$          & \multicolumn{1}{c|}{$26.3 \pm 4.5$}          & $61.6 \pm 1.4$          & \multicolumn{1}{c|}{$28.3 \pm 5.1$}          & $64.7 \pm 1.6$          \\
TaxoPrompt                                              & \multicolumn{1}{c|}{$\underline{51.9 \pm 6.3}$}    & $\underline{78.6 \pm 1.7}$    & \multicolumn{1}{c|}{ $\underline{58.3 \pm 3.8}$}    &  $\underline{78.1 \pm 0.7}$    & \multicolumn{1}{c|}{ $\underline{49.5 \pm 3.7}$}    & $74.4 \pm 1.4$          \\ \hline
\modelname\                                               & \multicolumn{1}{c|}{$\mathbf{63.4 \pm 1.9}$} & $\mathbf{85.1 \pm 0.3}$ & \multicolumn{1}{c|}{$\mathbf{63.2 \pm 4.1}$} & $\mathbf{82.5 \pm 1.2}$ & \multicolumn{1}{c|}{$\mathbf{52.7 \pm 4.1}$} & $\mathbf{78.1 \pm 1.5}$ \\ \hline
\end{tabular}}
%\vspace{-5mm}
\label{table:results}
\end{table}

\section{Experimental Results}
\label{sec:experiment_result}

The performance of the competing methods on three benchmarks is shown in Table \ref{table:results}. \modelname\ exhibits significant improvements over the baselines across all the metrics. 
We analyze the baseline performance below as they evolve chronologically. 

\begin{figure}[t]
\centering
\includegraphics[width=1.0\textwidth]{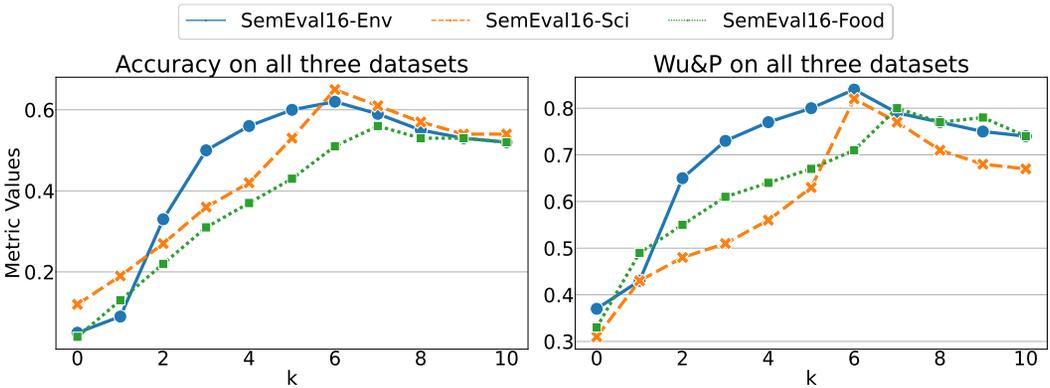}
\caption{Performance of \modelname\ for the different number of examples ($k$) included in the prompt.}
\label{fig:ab1}
%\vspace{-5mm}
\end{figure}

\begin{itemize}[leftmargin=*]
    \item BERT+MLP adopts a vanilla approach by merely leveraging the surface names of the terms while neglecting the inherent structural and lexical-syntactic information of the seed taxonomy. Furthermore, BERT fails to capture the contextual relationship among the terms, thereby leading to poor performance.
    
    \item First-generation baselines like TAXI incorporate information like contextual, lexical and semantic features for hypernymy detection, which boosts their performance. However, they do not leverage the powerful LMs or taxonomy structure; therefore, they cannot maintain the coherency of the taxonomy.
    
    \item Second-generation models, such as TaxoExpan and STEAM, incorporate lexico-syntactic features along with structural features like dependency paths and ego networks, improving their performance over first-generation methods. They, too, overlook the hierarchy of the taxonomy, thus failing to model the coherency of the taxonomy.
    
    \item Third-generation methods like TEMP employ taxonomy paths to learn the taxonomy structure using transformers, enhancing their performance. But they fall short of capturing the local neighbourhood intricacy. However, their counterparts, such as HEF, integrate ego networks to comprehend the local neighbourhood using ego networks alongside hierarchical taxonomy modeling, thereby outperforming TEMP. However, these approaches predominantly rely on PLMs and neglect to leverage the powerful LLMs or prompting techniques, limiting them to effectively modelling contextual features.
    
    \item Recent research on geometrical embeddings for taxonomy modeling also shows promising results. For instance, BoxTaxo models the hypernymy relationship using the enclosure condition of hyperrectangles \cite{abboud2020boxe}. However, despite its innovative approach, BoxTaxo neglects the comprehensive global structure of the taxonomy, leading to subpar performance.

    \item The enhanced performance demonstrated by prompt-based methods like TaxoPrompt confirms our conjecture regarding the expressive capabilities inherent in prompting methodologies. TaxoPrompt captures the taxonomy structure using a random walk while the local structure is learned via masked language modeling. Nevertheless, its performance is somewhat compromised since it relies on PLMs, which inherently struggle with modeling structural information from the prompts.

    \item \modelname\ improves on the third-generation models, which utilize the taxonomy hierarchy along with prompts by proposing a taxonomy-oriented prompt template, enhanced by the power of LLMs that captures the global structure and local nuances of the taxonomy. We propose a prompt engineering strategy which uses clustering techniques and BM25 ordering to craft domain-specific prompts. We enrich prompts with diverse global samples to incorporate global features. Experimental results indicate that \modelname\ effectively leverages the knowledge in the LLMs augmented with refined prompting strategies to exploit the hierarchical structure of the taxonomies.

\end{itemize}

We conduct a \textit{t}-test and calculate the corresponding \textit{p}-value, comparing the performance of the best baseline, TaxoPrompt, with that of the \modelname. We consider the test set of all the benchmarks for the experiment. The test set is divided into batches of size 10 and the precision of these batches is computed. With a critical value of 2.3060 for the degrees of freedom (8), the computed t-statistic of 3.88064 leads us to reject the null hypothesis $H_0$, suggesting that the mean performance of the two models differs significantly. This inference is further supported by the \textit{p}-value test, which yields a \textit{p}-value of 0.00467, well below the significance threshold of 0.01. Therefore, supported by both the statistical tests and empirical results, \modelname\ convincingly outperforms the best baseline, TaxoPrompt.

\begin{figure}[!t]
\centering
\includegraphics[width=1.0\textwidth]{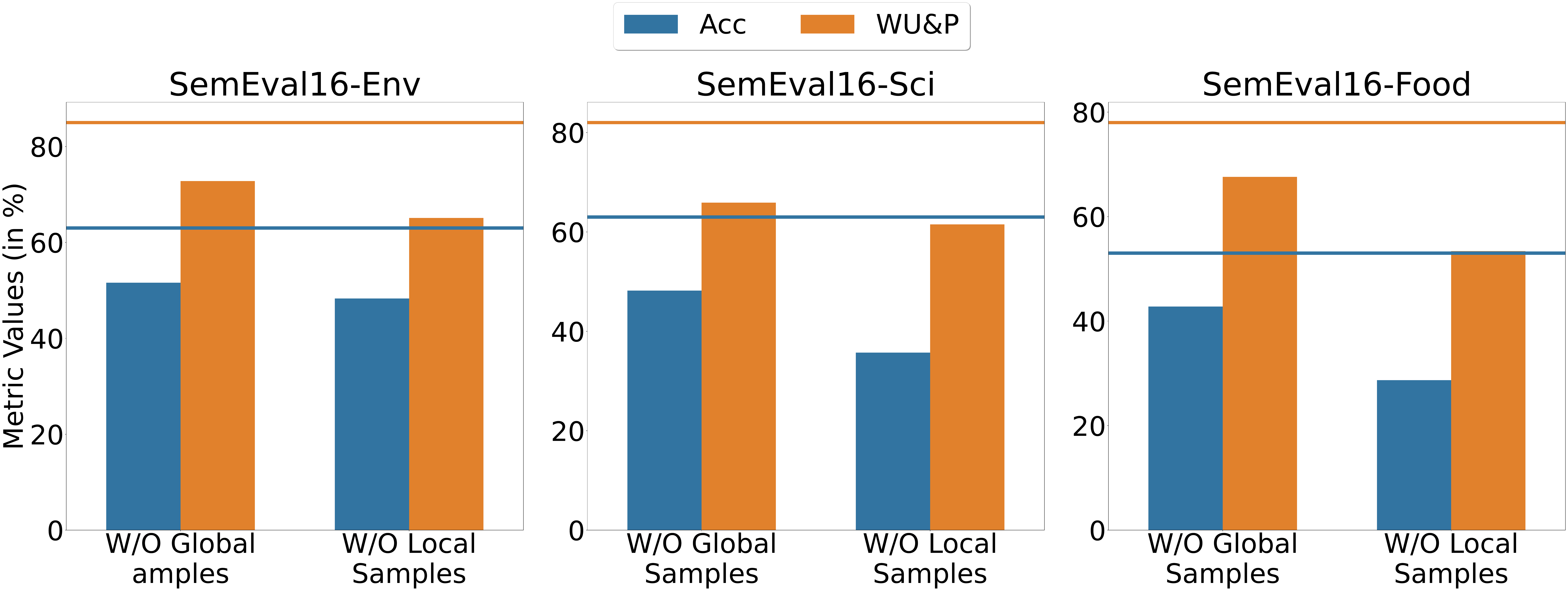}
\caption{Performance of \modelname\ without local and global samples included in the prompt. The horizontal lines represent the performance of \modelname\ when both global and local samples are included.}
\label{fig:glolo}
%\vspace{-5mm}
\end{figure}

\section{Ablation Studies}
\label{sec:ablation}
We conduct the following ablation studies to understand the efficacy of various components of \modelname: (i) the number of instances in the prompt and (ii) the effect of the global and local samples.

\subsection{The Number of Instances in the $K$-shot Prompt}

We study the effect of change in the value $k$ defined in Equation \ref{eq:prompt}. To study the effectiveness of the instances supporting the instruction in a $k$-shot prompt, we vary the number of instructions, denoted as $k$. When $k$ equals 0, the prompt is designated as "zero-shot". As shown in Fig. \ref{fig:ab1}, the zero-shot prompt yields the worst performance, confirming the failure of the zero-shot prompt to capture structural information of the seed taxonomy. Also, a zero-shot prompt, supplied with no examples, does not tell the model how an output is expected. Therefore, the model generates the hypernyms outside the set of anchor nodes.  In contrast, a one-shot prompt supplies an example to the model to learn how output is returned. Therefore, the performance improves. Further, as the value of $k$ increases (i.e., few-shot prompt), optimal performance is observed with a 7-shot prompt. However, as the value of $k$ increases, the performance plateaus and subsequently decreases, which shows that the model struggles to capture generic structural information as it succumbs to overfitting on the prompt instances. Therefore, an optimal value of $k$ is [5,7] as the model balances global overview and local intricacies without overfitting the limited training data.

\subsection{The Effect of Global and Local Samples}
The impact of both global and local samples on the prompt is shown in Figure \ref{fig:glolo}. The absence of either results in a degradation in model performance. Eliminating global samples leads to an average decrease of 21.6\% accuracy and 17.1\% Wu\&P score. This decline is attributed to the structural summary of the taxonomy encapsulated by the global samples; their removal simplifies the task to mere hypernymy detection. This decline is attributed to the structural summary of the taxonomy captured by the global samples; their removal reduces the task of taxonomy completion to mere hypernymy detection. However, removing local samples further exacerbates the performance, with accuracy dropping by an average of 38.3\% and Wu\&P score by 26.8\%. This underscores the importance of the local samples on the prompt as they capture the local neighbourhood around the query node.

\begin{table}[]
\caption{Examples showcasing the hypernyms predicted by \modelname\ and their comparisons with true hypernyms.}
\scalebox{0.99}{
\begin{tabular}{|l|l|l|l|}
\hline
\multicolumn{1}{|c|}{\textbf{Taxonomy}} & \multicolumn{1}{c|}{\textbf{Query Term $n_q$}} & \multicolumn{1}{c|}{\textbf{True Parent $n_p$}} & \multicolumn{1}{c|}{\textbf{Predicted Parent $n_p$}} \\ \hline
SemEval16-Env                           & soil resources                                 & natural resources                               & natural resources                                    \\ \hline
SemEval16-Env                           & biological standard                            & environmental standard                          & environmental law                                    \\ \hline
SemEval16-Env                           & groundwater                                    & water                                           & water resources                                      \\ \hline
SemEval16-Env                           & radioactive effluent                           & radioactive waste                               & radioactive waste                                    \\ \hline
SemEval16-Env                           & inflammable product                            & dangerous substance                             & dangerous substance                                  \\ \hline
SemEval16-Env                           & reptile                                        & animal life                                     & wildlife                                             \\ \hline
SemEval16-Sci                           & epigraphy                                      & paleography                                     & archaeology                                          \\ \hline
SemEval16-Sci                           & sociolinguistics                               & linguistics                                     & linguistics                                          \\ \hline
SemEval16-Sci                           & dental anatomy                                 & gross anatomy                                   & anatomy                                              \\ \hline
SemEval16-Sci                           & pteridology                                    & botany                                          & botany                                               \\ \hline
SemEval16-Food                          & choline                                        & b complex vitamin                               & vitamin                                              \\ \hline
SemEval16-Food                          & potage                                         & soup                                            & soup                                                 \\ \hline
SemEval16-Food                          & mozzarella                                     & cheese                                          & cheese                                               \\ \hline
SemEval16-Food                          & sauterne                                       & white wine                                      & liqour                                               \\ \hline
\end{tabular}
}
%\vspace{-5mm}
\end{table}

\section{Case Studies and Error Analysis}
\label{sec:case}
In this section, we conduct case studies and perform error analysis to demonstrate the efficacy of the \modelname\ framework, achieved through an assessment of predictions across various query terms within the SemEval16-Env, SemEval16-Sci, and SemEval16-Food datasets. Notably, for unambiguous, straightforward and widely understood query concepts such as \textit{soil resources}, \textit{radioactive effluent}, \textit{sociolinguistics} and \textit{mozzarella}, \modelname\ accurately retrieves appropriate hypernyms as their definitions are self-sufficient in capturing their semantics. However, with query terms like \textit{biological standard}, \textit{groundwater}, \textit{reptile} and \textit{choline}, the model often returns incorrect hypernyms, although they are similar to the true hypernym semantically. This is due to the fact that the model gets confused with the hypernyms' meaning, creating ambiguity as sometimes definitions might not be sufficient to represent the real meaning. This issue is particularly pronounced in the SemEval16-Food dataset, as illustrated in Table \ref{table:results}, where numerous concepts refer to exquisite gourmet food items whose precise meanings might not be adequately captured by their surface names or definitions. For instance, \textit{choline} is a \textit{b complex vitamin}, but its definition has no mention of any word close to vitamin. Therefore, the model attaches it to \textit{vitamin}, semantically closer to the true parent. Thus, thorough analysis identifies a recurring scenario wherein \modelname\ struggles to perform effectively, mainly when the description fails to represent the term's true semantics adequately. In such instances, the learnt parameters cannot accurately capture the hypernymy relationships between query terms and parent nodes.

\section{Conclusion}
\label{sec:conclusion}
We propose \modelname, a novel taxonomy expansion framework that leverages a seed taxonomy as self-supervision data to capture both its global coherency and local intricacies through a few-shot prompt. Integrating $k$-shot prompt and LLM helps overcome the low-resource problems faced by previous methods. We generate diverse global sample pools to encapsulate the structural essence of the taxonomy alongside local sample pools that delineate the local neighbourhood around a node, thereby enhancing the comprehension for hypernymy detection. These pools help in creating a robust $k$-shot prompt. Additionally, the anchor nodes from the seed taxonomy that are incorporated into the prompt for retrieval also provide a better structural overview. We fine-tune the low-rank parameters of the LLM using the proximal policy optimizer to align the generation with true hypernyms and restrict LLM from generating extraneous tokens, thereby reducing the inference time. Extensive experiments on three real-world datasets -- SemEval16-Env, SemEval16-Sci and SemEval16-Food demonstrate the efficacy of \modelname\ framework on taxonomy expansion task, surpassing the state-of-the-art methods. Ablation studies reveal insights into the optimal number of examples utilized in the few-shot prompting strategy and underscore the significance of incorporating both global and local samples into the prompt generation process. Furthermore, case studies and error analyses demonstrate the pivotal role of the definitions in the performance of \modelname, as it excels when definitions are unambiguous. Therefore, future works are needed to leverage LLMs in conjunction with lexico-syntactic reward functions to fine-tune LLMs, aligning the generation of definitions with the actual meaning of terms, thus disambiguating the semantics of terms within the taxonomy. Furthermore, subsequent efforts may focus on taxonomy augmentation through prompting, involving the addition of novel relations for which the hypernyms are not present in the seed taxonomy.

\bibliographystyle{ACM-Reference-Format}
\bibliography{bibliography}

\clearpage

\appendix

\tikzstyle{startstop} = [rectangle, rounded corners, 
minimum width=3cm, 
minimum height=1cm,
text centered, 
text width = 5cm,
draw=black, 
fill=red!30]

\tikzstyle{io} = [trapezium, 
trapezium stretches=true, % A later addition
trapezium left angle=70, 
trapezium right angle=110, 
minimum width=3cm, 
minimum height=1cm, text centered, 
draw=black, fill=blue!30]

\tikzstyle{process} = [rectangle, 
minimum width=3cm, 
minimum height=1cm,  
text width=15cm, 
draw=black, 
fill=green!30]

\tikzstyle{process2} = [rectangle, 
minimum width=3cm, 
minimum height=1cm,  
text width=10cm, 
draw=black, 
fill=green!30]

\tikzstyle{cot} = [rectangle, 
minimum width=3cm, 
minimum height=1cm,  
text width=10cm, 
draw=black, 
fill=orange!30]

\tikzstyle{decision} = [diamond, 
minimum width=3cm, 
minimum height=1cm, 
text centered, 
draw=black, 
fill=green!30]

\tikzstyle{arrow} = [thick,->,>=stealth]

\newtcolorbox{mybox1}[1][]{boxsep=1pt,left=1pt,right=1pt,colback=red!5!white,colframe=red!75!black,fonttitle=\bfseries,title=#1}
\newtcolorbox{mybox2}[1][]{boxsep=1pt,left=1pt,right=1pt,colback=pink!5!white,colframe=pink!75!black,fonttitle=\bfseries,title=#1}
\newtcolorbox{mybox3}[1][]{boxsep=1pt,left=1pt,right=1pt,colback=purple!5!white,colframe=purple!75!black,fonttitle=\bfseries,title=#1}
\newtcolorbox{boxA}{boxsep=1pt,left=1pt,right=1pt,colframe=gray!80}

\section{Prompt-Engineered Dataset Example}
\label{app:prompt}

This section delves into the prompt template formulated as outlined in Section \ref{subsec:prompt}. An example of a data sample from the evaluation dataset given below contains the following prompt $p(n_q)$ pertaining to the query node \textit{evaluation of resources} within the environment taxonomy. The candidate hypernym list $\mathcal{N}^0$, distinguished by its \textbf{\textit{cyan}} hue, is a part of the instruction set $\mathcal{I}$. Moreover,  terms highlighted in \textit{\textbf{green}} represent local samples $s_l \in S_l$ specific to the query term \textit{evaluation of resources}, whereas those highlighted in \textit{\textbf{yellow}} signify the global samples $s_g \in S_g$. The context represents the definitions of the surface names ($d_g, d_l$ and $d_q$) of the hyponyms in the prompt, thereby representing a comprehensive summary to understand the nuances of the taxonomy.
\\
\begin{center}
\begin{mybox3}[$P(n_q)$ for \textit{evaluation of resources} in \textbf{Environment} Taxonomy] \small 
You are an assistant to hypernym prediction and sorting.\\
Given a term, its context and a list of candidate hypernym answers to this term, You need to rank these candidate terms in the list to let candidate answers which are most possible to be the hypernym or parent term to the given term and return the list.\\
\colorbox{cyan}{Candidate Hypernym List} = [polluted area, underwater mineral resources, animal life, environment, EU environmental policy, $\dots$]\\
A few examples of hypernym-hyponyms are given as:\\
\colorbox{green}{TERM: management of resources}\\
CONTEXT: management of resources: the deliberate and strategic use of resources (natural, financial, human, etc.) to achieve specific objectives, often involving conservation and sustainable practices.\\
HYPERNYM: environmental policy\\
\colorbox{green}{TERM: replacement of resources}\\
CONTEXT: replacement of resources: the process of replenishing natural resources that have been consumed or depleted, often through sustainable practices like reforestation or renewable energy sources.\\
HYPERNYM: management of resources\\
\colorbox{yellow}{TERM: natural disaster}\\
CONTEXT: natural disaster: a natural disaster is a severe and sudden event caused by environmental factors that can result in significant damage and loss of life. examples include earthquakes, floods, hurricanes, tsunamis, and volcanic eruptions.\\
HYPERNYM: degradation of the environment\\
\colorbox{yellow}{TERM: toxic substance}\\
CONTEXT: toxic substance: a toxic substance is a chemical or compound that can cause harm or illness to living organisms, either through ingestion, inhalation, or contact with skin.
HYPERNYM: dangerous substance\\
\colorbox{yellow}{TERM: bird}\\
CONTEXT: bird: birds are a group of warm-blooded vertebrates constituting the class aves, characterized by feathers, toothless beaked jaws, the laying of hard-shelled eggs, a high metabolic rate, and a lightweight but strong skeleton.\\
HYPERNYM: animal life\\
TERM: evaluation of resources\\
CONTEXT: evaluation of resources: the process of assessing the availability, quality, and economic value of natural resources for sustainable management and utilization.\\
HYPERNYM:
\end{mybox3}
\label{center:prompt}
\end{center}

\clearpage

\section{Data Augmentation}
\label{app:aug}
While prompting an LLM using the template specified in Appendix \ref{app:prompt}, in addition to predicting hypernyms, the LLM also generates novel relations. These relations offer the potential for enhancing the taxonomy further. However, assessing these relations becomes arduous when the corresponding hypernyms are absent from the taxonomy. To illustrate the process of hypernym prediction and relation augmentation, consider the following example from SemEval16-Env with the input prompt highlighted in \textit{blue}. The prompt yields hypernym \textit{land degradation} for the term \textit{desertification}. Subsequently, a set of relations are predicted, such as \textit{acidification} $\rightarrow$ \textit{environmental pollution}. However, terms such as "environmental pollution" and "land use change" are not present in the seed taxonomy. Therefore, the incorporation of these terms proves infeasible and hampers the evaluation.

\begin{center}
\begin{mybox3}[Prompt with output for \textit{desertification} in \textbf{Environment} Taxonomy]
\textcolor{blue}{You are an assistant to hypernym prediction and sorting.\\
Given a term, its context and a list of candidate hypernym answers to this term, You need to rank these candidate terms in the list to let candidate answers which are most possible to be the hypernym or parent term to the given term and return the list.\\
Candidate Hypernym List = [polluted area, underwater mineral resources, animal life, environment, $\dots$]\\
A few examples of hypernym-hyponyms are given as:\\
TERM: soil pollution\\
CONTEXT: soil pollution: the contamination of soil with harmful substances $\dots$\\
HYPERNYM: pollution\\
TERM: pollution from agricultural sources\\
CONTEXT: pollution from agricultural sources: contamination of the  $\dots$\\
HYPERNYM: pollution\\
TERM: waste\\
CONTEXT: waste: waste refers to materials that are discarded because they are  $\dots$\\
HYPERNYM: environment\\
TERM: biodiversity\\
CONTEXT: biodiversity: this is the variety of life in a particular it $\dots$\\
HYPERNYM: biosphere\\
TERM: desertification\\
CONTEXT: desertification: desertification is the process by which fertile $\dots$\\
HYPERNYM:} land degradation\\
TERM: acidification\\
CONTEXT: acidification: the lowering of the pH of a solution, often due to the $\dots$\\
HYPERNYM: environmental pollution\\
TERM: marine pollution\\
CONTEXT: marine pollution: pollution of the ocean or other bodies of saltwater  $\dots$\\
HYPERNYM: water pollution
TERM: air pollution\\
CONTEXT: air pollution: the presence of pollutants in the air that can $\dots$\\
HYPERNYM: environmental pollution\\
TERM: deforestation\\
CONTEXT: deforestation: the removal of forests and trees, often for agricultural $dots$\\
HYPERNYM: land use change
\end{mybox3}
\end{center}

\clearpage

\section{Reward Computation}
\label{app:reward}

During the training of \modelname\ using PPO as mentioned in Section \ref{sec:llm}, we defined the reward functions aimed at aligning the predicted hypernym with true hypernym in Section \ref{subsubsec:reward}. In this context, we describe the computation of rewards through illustrative examples.

\subsection{Label Reliability}
Label reliability involves verifying whether the predicted hypernym is congruent with the true hypernym. Here, \textit{mineral} is different from \textit{mineral resources}. Therefore, the reward assigned in this scenario amounts to zero. 
\begin{boxA}
$n_p$: Mineral Resources\\
$\hat{n}_p$: Mineral\\
$R_m=I(n_p = \hat{n}_p)=0$

\end{boxA}

\subsection{Semantic Consistency}
Semantic coherency is measured by computing the cosine similarity between the predicted and true hypernyms using their vector representations derived from their surface names. For instance, \textit{Environmental Standard} bears a striking semantic resemblance \textit{Environmental Law}, yielding a semantic similarity score of 0.60625.
\begin{boxA}
$n_p$: Environmental Standard\\
$\hat{n}_p$: Environmental Law\\
$R_c= \text{cos-sim}(e, \hat{e})=0.60625$

\end{boxA}

\subsection{Label Length Conformity}
Label length conformity evaluates the variance between lengths of true hypernym and predicted hypernym. Further, this variance is normalized by the sum of their lengths. If a difference is observed, it is negated and passed as a penalty.  
\begin{boxA}
$n_p$: b complex vitamin\\
$\hat{n}_p$: vitamin\\
$\mathcal{L}_{n_p}= 17$\\
$\mathcal{L}_{\hat{n}_p}=7$\\
$\big|\mathcal{L}_{\hat{n}_p}-\mathcal{L}_{n_p}\big|=\big|7-17\big|=10$\\
$\therefore -\frac{\big|\mathcal{L}_{\hat{n}_p}-\mathcal{L}_{n_p}\big|}{\mathcal{L}_{\hat{n}_p}+\mathcal{L}_{n_p}}=-\frac{\big|7-17\big|}{\big|7+17\big|}=-\frac{5}{12}=-0.42$
\end{boxA}

\subsection{Token Count Consistency}
Token count consistency assesses the number of unique common tokens between the true hypernym and the predicted hypernym. It is subsequently normalized by the total count of unique tokens found in both entities collectively. Consequently, for the terms \textit{exploitation of resources} and \textit{over-exploitation of resources}, the common token count stands at 3 out of a combined total of 4 unique tokens.
\begin{boxA}
$n_p$: Exploitation of Resources\\
$\hat{n}_p$: Over-exploitation of Resources\\
$\mathcal{S}_{n_p}=$\{"Exploitation", "of", "Resources"\}\\
$\mathcal{S}_{\hat{n}_p}=$\{"Over", "Exploitation", "of", "Resources"\}\\
$\big|\mathcal{S}_{\hat{n}_p}\cap\mathcal{S}_{n_p}\big|$=3\\
$\therefore \frac{\big|\mathcal{S}_{\hat{n}_p}\cap\mathcal{S}_{n_p}\big|}{\big|\mathcal{S}_{\hat{n}_p}\cup\mathcal{S}_{n_p}\big|}=\frac{3}{4}=0.75$
\end{boxA}

\subsection{Fuzzy Label Matching}
Fuzzy label matching involves comparing generated hypernyms with true hypernyms through Levenshtein distance-based similarity assessment. It computes four different ratios to measure the deviation of the predicted hypernym from the true hypernym. The four ratios are computed on the pairs of predicted and true hypernyms below.

Edit distance ratio measures the Levenshtein distance-based similarity between two strings. It helps align the complete predicted hypernym with the true hypernym. 
\begin{boxA}
\textbf{Edit Distance Ratio}\\
$n_p$: b complex vitamin\\
$\hat{n}_p$: vitamin b\\
$\mathcal{L}_{\hat{n}_p}+\mathcal{L}_{n_p}=17 + 9 = 26$\\
$\text{Levenshtein-distance}(n_p, \hat{n}_p)=12$\\
$F_r= \text{edit-distance-ratio}(n_p, \hat{n}_p)= \frac{26-12}{26}=0.538$
\end{boxA}

Partial edit distance ratio computes the Levenshtein distance similarity ratio on the substrings. It selects the shortest substring and matches it with all the substrings of the same length. 
\begin{boxA}
\textbf{Partial Edit Distance Ratio}\\
$n_p$: b complex vitamin\\
$\hat{n}_p$: vitamin b\\
$\text{shorter string} = \hat{n}_p$\\
$\text{substring}_{n_p} = \text{vitamin}$\\
$\mathcal{L}_{\text{substring}_{n_p}}+\mathcal{L}_{\hat{n}_p}=7 + 9 = 16$\\
$\text{Levenshtein-distance}(\text{substring}_{n_p}, \hat{n}_p)=2$\\
$F_{pr}= \text{edit-distance-ratio}(\text{substring}_{n_p}, \hat{n}_p)= \frac{16-2}{16}=0.875$
\end{boxA}

Token sort ratio tokenizes the strings, sorts them alphabetically, concatenates them together and then computes the Levenshtein edit distance ratio between them.
\begin{boxA}
\textbf{Token Sort Ratio}\\
$n_p$: b complex vitamin\\
$\hat{n}_p$: vitamin b\\
$\text{sorted}_{n_p} = \text{b complex vitamin}$\\
$\text{sorted}_{\hat{n}_p} = \text{b vitamin}$\\
$\mathcal{L}_{\text{sorted}_{\hat{n}_p}}+\mathcal{L}_{\text{sorted}_{n_p}}=17 + 9 = 26$\\
$\text{Levenshtein-distance}(\mathcal{L}_{\text{sorted}_{\hat{n}_p}},\mathcal{L}_{\text{sorted}_{n_p}})=8$\\
$F_{tsor}= \text{edit-distance-ratio}(\mathcal{L}_{\text{sorted}_{\hat{n}_p}}+\mathcal{L}_{\text{sorted}_{n_p}})= \frac{26-8}{26}=0.692$
\end{boxA}

The token set ratio operates similarly to the token sort ratio, except it computes the edit distance ratio on the common tokens of both strings. Its objective is to mitigate discrepancies within the strings. Specifically, it performs the following operations.
\begin{itemize}
    \item Intersection of string one with string two.
    \item Difference of set of tokens of string one from the set of tokens of string two.
    \item Difference of set of tokens of string two from the set of tokens of string one.
\end{itemize}
Subsequently, it merges difference sets with the intersection set and then computes the edit distance ratio of the intersection set with these two merged difference sets and also the ratio between the merged difference sets. Here, we discuss an example of the maximum edit distance only.
\begin{boxA}
\textbf{Token Set Ratio}\\
$n_p$: b complex vitamin\\
$\hat{n}_p$: vitamin b\\
$\text{sorted}_{n_p} = \text{b complex vitamin}$\\
$\text{sorted}_{\hat{n}_p} = \text{b vitamin}$\\
$\text{intersection} = \text{b vitamin}$\\
$\text{diff1} = \text{complex}$\\
$\text{diff2} = \text{\{\}}$\\
$\text{merge1} = \text{b complex vitamin}$\\
$\text{merge2} = \text{b vitamin}$\\
$F_{tser}= \text{max-edit-distance-ratio}= 
\text{edit-distance-ratio}(\text{intersection}, \text{merge2})=\frac{18-0}{18}=1.0$
\end{boxA}
   
\end{document}